\documentclass{article}

\usepackage{microtype}
\usepackage{graphicx}
\usepackage{subfigure}
\usepackage{booktabs} 
\usepackage{amsmath}
\usepackage{breqn}
\usepackage{float}
\usepackage{caption}

\usepackage{hyperref}

\usepackage[accepted]{icml2020}

\icmltitlerunning{Investigating Saturation Effects in Integrated Gradients}

\begin{document}

\twocolumn[
\icmltitle{Investigating Saturation Effects in Integrated Gradients}



\icmlsetsymbol{equal}{*}

\begin{icmlauthorlist}
\icmlauthor{Vivek Miglani}{fb}
\icmlauthor{Narine Kokhlikyan}{fb}
\icmlauthor{Bilal Alsallakh}{fb}
\icmlauthor{Miguel Martin}{fb}
\icmlauthor{Orion Reblitz-Richardson}{fb}
\end{icmlauthorlist}

\icmlaffiliation{fb}{Facebook AI}

\icmlcorrespondingauthor{Vivek Miglani}{vivekm@fb.com}

\icmlkeywords{Machine Learning, ICML}

\vskip 0.3in
]



\printAffiliationsAndNotice{} 

\begin{abstract}
Integrated Gradients has become a popular method for post-hoc model interpretability. Despite its popularity, the composition and relative impact of different regions of the integral path are not well understood. We explore these effects and find that gradients in saturated regions of this path, where model output changes minimally, contribute disproportionately to the computed attribution. We propose a variant of Integrated Gradients which primarily captures gradients in unsaturated regions and evaluate this method on ImageNet classification networks. We find that this attribution technique shows higher model faithfulness and lower sensitivity to noise compared with standard Integrated Gradients. 
A notebook illustrating our computations and results is available at \href{https://github.com/vivekmig/captum-1/tree/ExpandedIG}{https://github.com/vivekmig/captum-1/tree/ExpandedIG}.

\end{abstract}

\section{Introduction}
\label{submission}

The need for interpretability has grown substantially in the past years, driven by the promise of neural networks in a variety of fields.
In many application areas, such as health-care, finance, and several scientific fields, black-box predictions are of limited usefulness if they are not coupled with human-understandable explanations.
Many methods for interpreting the decisions of neural networks have been proposed including DeepLift \cite{shrikumar2017learning} and Integrated Gradients \cite{sundararajan2017axiomatic}. These methods assign a scalar value to each input quantifying its importance to the model's output. Quantitative evaluation of these  methods is challenging due to the lack of a ground truth for input importance to a particular model.

Integrated Gradients (IG) has gained significant attention since it satisfies desirable axiomatic properties such as completeness, sensitivity, and linearity and is shown to be the unique method satisfying these properties \cite{sundararajan2017axiomatic}. Instead of relying on the gradients at the input, where a model output is often saturated and does not change significantly, IG sums gradients over gradual modification of the input from a  baseline value to the original value of the input.
This integral path is controlled by a value, $\alpha$, called the scaling factor.
The model output changes substantially along this path, gradually leading to improved attribution~\cite{sundararajan2016gradients}.

In this work, we delve deeper into IG and quantitatively measure how the saturation of model output with respect to  $\alpha$ impacts overall attribution. Our main contributions are:
\begin{itemize}
\item We compare the contribution of gradients in saturated regions of $\alpha$, where the model output changes minimally, to those of unsaturated regions, where the model output changes substantially. We find that the gradients in saturated regions can have a large impact on the computed attribution.
\item Based on this observation, we propose a variant of IG, we call Split IG, which restricts the integral to regions where the model output changes substantially. 
\item We evaluate Split IG both qualitatively and quantitatively for various models trained on ImageNet. We find that Split IG exhibits better fidelity to the model and less sensitivity to small input perturbations compared with standard IG.
\end{itemize}

\section{Integrated Gradients (IG)}

IG attributes a network's prediction to its input features by integrating gradients along a straight-line path between a baseline input and the original input \cite{sundararajan2017axiomatic}. Formally, the $i^{th}$ index of integrated gradients can be defined for a continuous function $F: R^n \rightarrow R$, input $x$ and baseline $x'$ as follows:
\[
IG_i(x, x') = (x_i - x_i') \cdot \int_0^1 \frac{\partial F(x_i' + \alpha\cdot(x_i - x_i'))}{\partial x_i} d \alpha
\]
IG has been shown to satisfy desirable axiomatic properties including completeness. This property ensures that the attributions sum to the difference between the model output computed respectively at the baseline and at the input, essentially distributing the total change in model output across gradual changes of the input. The completeness property is a direct result of the fundamental theorem of calculus for path integrals and can be stated as follows:
\[\sum_{i=1}^n IG_i(x, x') = F(x) - F(x')\]
In addition to the axiomatic properties of IG, it has also been suggested as a solution to the problem of saturation, the flattening of the model's output around the input. 


Figure~\ref{fig:concept}a shows the logit of a target class for a sample image as $\alpha$ changes from 0 (grey baseline) to 1 (original image), computed for an Inception-v3 model trained on ImageNet. We use this logit as the target function $F$ in IG.
We note that the substantial increase in $F$ occurs for $\alpha \in [0, 0.33]$, after which $F$ remains nearly constant as $\alpha$ scales up to 1. The simple approach of Saliency \cite{simonyan2013deep} takes the gradient at $\alpha = 1$ as a measure of attribution.
However, this point is in the saturated region where $F$ is not changing substantially with respect to $\alpha$. In contrast, IG adds up gradients along the entire path, capturing regions where $F$  changes substantially. This enables IG to capture how the model changes its prediction for a given target class from a baseline value to the final logit, potentially leading to improved attribution.

In this work, we investigate the aforementioned intuition about IG to (1) validate whether gradients in the unsaturated regions along the integral path dominate the computation and (2) to evaluate how the gradients in the saturated region affect the computed attribution.

\section{Split Integrated Gradients}
To separate the contribution of saturated areas from unsaturated areas along the integral path, we propose Split Integrated Gradients as a variant of IG.
Split IG  divides the integral into two parts. The first part corresponds to the substantial increase in $F$, while the second part corresponds to the saturated region where $F$ does not change substantially.

\begin{figure}[ht]
    \centering
    \includegraphics[width=0.99\linewidth]{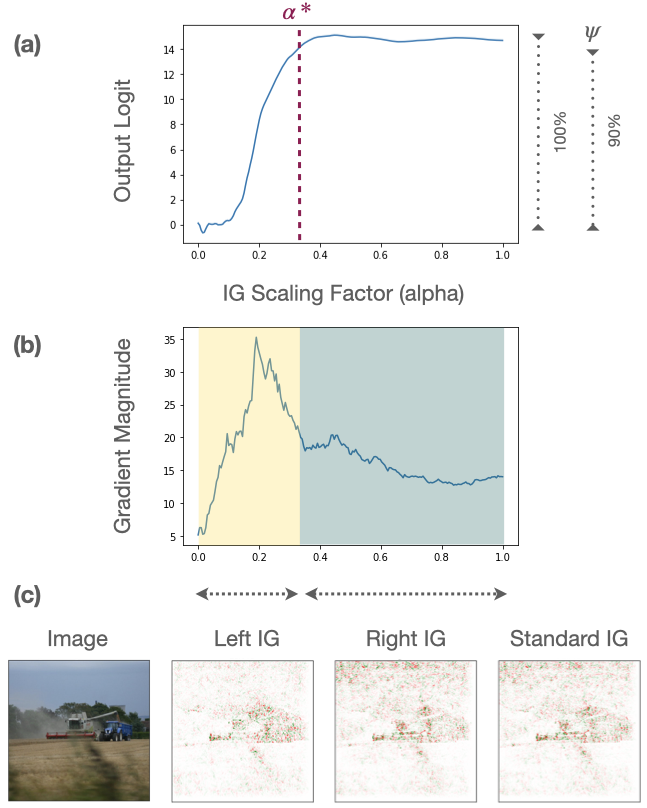}
    \caption{\textbf{Overview of our method.}
    (a) Output logit for ImageNet class \texttt{harvester} in Inception-v3 as the input changes from a grey baseline ($\alpha = 0$) to the original image ($\alpha = 1$), depicted in (c).
    The dotted line marks $\alpha^*$ for $\psi = 0.9$.
    (b) The magnitude of the computed gradient over $\alpha$.
    (c) Restricting the integral path when computing attribution for an input image to $[0, \alpha^*]$ (Left IG) and to  $[\alpha^*, 1]$ (Right IG), compared with standard IG.
    }
    \label{fig:concept}
\end{figure}

For a given percentage threshold such as $\psi = 90\%$, we first determine the minimum value of $\alpha$ such that the target output $F(x' + \alpha \cdot (x - x'))$ exceeds $\psi$
of the effective output range along the integral path. Formally, we define this threshold as follows:
\begin{multline}
\label{ineq}
\alpha^* = \inf \lbrace \alpha \in [0,1] \textnormal{ s. t. } \\ F(x' +  \alpha \cdot (x - x')) > F(x') + \psi \cdot (F(x) - F(x')) \rbrace
\end{multline}

After determining $\alpha^*$, we divide the IG integral into two parts split at $\alpha^*$, as illustrated in Figure~\ref{fig:concept}b: 
\[LeftIG_i(x, x') = (x_i - x_i') \int_0^{\alpha^*} \frac{\partial F(x_i' + \alpha \cdot (x_i - x_i'))}{\partial x_i} d \alpha\]
\[RightIG_i(x, x') = (x_i - x_i') \int_{\alpha^*}^1 \frac{\partial F(x_i' + \alpha \cdot (x_i - x_i'))}{\partial x_i} d \alpha\]
Since $\alpha^* \in [0, 1]$, it is clear that the sum of these two integrals is equal to the full integrated gradients:
\[LeftIG_i(x, x') + RightIG_i(x, x') = IG_i(x, x')\]
Furthermore, assuming  $F$ is continuous, the two sides of the inequality in Eq~\ref{ineq} become equal at $\alpha^*$: \[F(x' + \alpha^*\cdot (x-x')) = F(x') + \psi \cdot (F(x) - F(x'))\]

Applying the completeness property to each of these integrals, we also find that:
\[\sum_{i=1}^n LeftIG_i(x, x') = \psi \cdot (F(x) - F(x'))\]
\[\sum_{i=1}^n RightIG_i(x, x') = (1 - \psi) \cdot (F(x) - F(x'))\]

The sum of the attributions of each of $LeftIG_i$ and $RightIG_i$ is determined by $\psi$. If $\psi$ is close to 1, the sum of $RightIG_i$ is usually small. Note that this does not directly imply that $RightIG$ does not contribute substantially to the computed IG; despite having a sum close to 0, the \textit{magnitude} of the gradients can be large as illustrated in Figure~\ref{fig:concept}b.

To measure the contribution of the saturated regions, we evaluate the ratio between $||RightIG||$ to $||LeftIG||$ for different thresholds of $\psi$, using both $\ell_1$ and $\ell_2$ norms. We conduct these experiments using different ImageNet models including Inception-v3~\cite{simonyan2014very}, ResNet-50~\cite{he2016deep}, and VGG-19~\cite{szegedy2016rethinking}. All experiments are performed with a 0 baseline, and the integrated gradients are computed using a right Riemann sum with 200 approximation steps. 
\begin{table}[H]
\centering
\caption{\textbf{Average ratio of norms (Right IG / Left IG)}. The norm of RightIG often exceeds that of LeftIG across model architectures and $\psi$ thresholds, showing that saturated regions have large contributions to IG.}
\vspace*{2mm}
\label{table:RelNorms}
\begin{tabular}{ c||c|c|c  }
 \multicolumn{4}{c}{Ratio between L2 norms } \\
 Model & $\psi = 0.9$ & $\psi = 0.95$ & $\psi = 0.99$\\
 \hline
 Inception-v3   & 2.4843    & 2.0840 &   1.5706\\
 ResNet-50 &   1.7684  & 1.4467   & 1.0505\\
 VGG-19 & 1.7262 & 1.2183 &  0.7604\\
\end{tabular}
\begin{tabular}{ c||c|c|c }
 \multicolumn{4}{c}{} \\
 \multicolumn{4}{c}{Ratio between L1 norms} \\
 Model & $\psi = 0.9$ & $\psi = 0.95$ & $\psi = 0.99$\\
 \hline
 Inception-v3   & 2.5371    & 2.1291 & 1.6021\\
 ResNet-50 &   1.8207  & 1.4896   & 1.0711\\
 VGG-19 & 1.8221 & 1.2772 &  0.7958\\
\end{tabular}
\end{table}
Table \ref{table:RelNorms} lists the relative norms averaged over 2500 samples from ImageNet validation set. We compute the attribution with respect to the logit of the labeled class. It is evident that RightIG contributes substantially to the overall integrated gradients, with the magnitude of the gradients often exceeding that of LeftIG, even with a threshold of 0.99. With a threshold of 0.90, the average ratio for Inception-v3 is approximately 2.5, suggesting that the contribution of right IG, where model output is highly saturated, is substantially larger than the contribution of left IG. This result counters the intuition of Integrated Gradients primarily capturing gradients along the path where model output changes substantially.
This, in turn, suggests that IG attributions can be significantly impacted by gradients in saturated regions.

Analyzing the L2 norm of the gradient as a function of $\alpha$ explains this effect, as shown in Figure~\ref{fig:concept}b. This magnitude spikes where the logit substantially increases.
Nevertheless, it remains substantially larger than zero in the saturated area where the logit function is approximately constant. 
\begin{table}[H]
\centering
\caption{\textbf{Average threshold $\alpha^*$ under different settings}. The value is often smaller than 0.5, suggesting that a large range of $alpha \in [0,1]$ is saturated.}
\vspace*{2mm}
\label{table:AvgThreshold}
\begin{tabular}{ c||c|c|c  }
 Model & $\psi = 0.9$ & $\psi = 0.95$ & $\psi = 0.99$\\
 \hline
 Inception-v3   & 0.280    & 0.355 & 0.493\\
 ResNet-50 & 0.382  & 0.464   & 0.606\\
 VGG-19 & 0.489 & 0.602 &  0.759\\
\end{tabular}
\end{table}
We compute $\alpha^*$ for each model and show the computed values in Table 3. The average threshold is generally smaller than $0.5$, implying that the saturated region of $\alpha$ can be larger than the unsaturated region. This explains why the overall magnitude of right IG often exceeds that of left IG, despite the individual gradients having smaller magnitude. 

\subsection{Evaluating Effects of Saturation}
A natural follow-up question to the results presented in the previous section revolves around understanding whether attributions computed from the saturated regions (RightIG) are similar to those of the unsaturated regions (LeftIG). Subsequently, we want to examine whether considering only the unsaturated regions when computing IG could improve attribution quality. We first qualitatively evaluate the differences between these attributions. 

\begin{figure}[ht]
    \centering
    \includegraphics[width=1\linewidth]{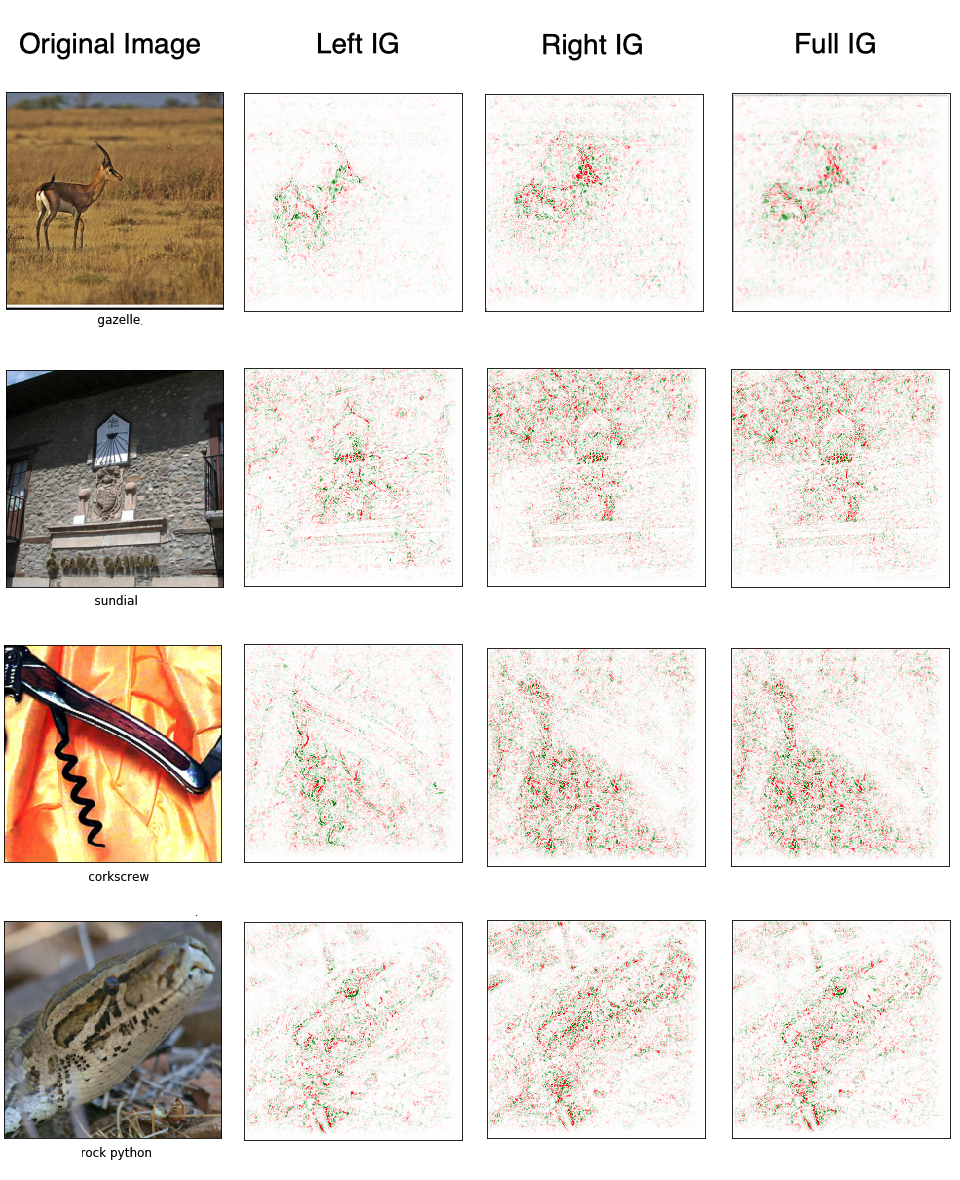}
    \caption{\textbf{Left IG, right IG, and full IG attributions generated using Captum~\cite{kokhlikyan2019pytorch} for randomly-selected ImageNet samples, classified using Inception-v3}. Full IG and Right IG generally result in similar attribution, while left IG attributions appear less noisy and more focused on salient features.
    Green and red resp. indicate positive and negative attribution.
    }
    \label{fig:sample_ims}
\end{figure}

In Figure~\ref{fig:sample_ims}, we demonstrate a few examples of Left IG, Right IG, and Full (original) IG attribution maps for Inception-v3 with $\psi = 0.9$. We provide further examples in the appendix. From these examples, differences are apparent between LeftIG and RightIG, with RightIG appearing substantially more noisy. This suggests that the gradient direction can shift substantially between saturated and unsaturated regions. We can further measure this effect quantitatively by computing the cosine similarity between these attributions. Table \ref{table:CosineSum} shows cosine similarity results averaged over 2500 samples from ImageNet validation set. 

\begin{table}[H]
\caption{\textbf{Average Cosine Similarity between left IG, right IG, and original (full) IG attributions for $\psi = 0.9$}. Full IG shows higher similarity to right IG than left IG, reinforcing that saturated gradients dominate. Similarity between left IG and right IG is consistently lower.}
\vspace*{2mm}
\label{table:CosineSum}
\begin{tabular}{ c||p{1.6cm}|p{1.6cm}|p{1.7cm} }
 Model & LeftIG $\leftrightarrow$ RightIG & LeftIG $\leftrightarrow$ Full IG & RightIG $\leftrightarrow$ Full IG\\
 \hline
 Inception-v3   &  0.2590    & 0.6265 &   0.8788\\
 ResNet-50 &   0.3495  & 0.7298   & 0.8522\\
 VGG-19 & 0.4686 & 0.7958 &  0.8667 \\
\end{tabular}
\end{table}

From these results, it is evident that integrated gradients for the saturated regions and unsaturated regions are substantially different, exhibiting low cosine similarity. RightIG has consistently higher cosine similarity to full (standard) IG, compared with LeftIG. This provides additional evidence that that saturated gradients often dominate the overall integrated gradients. 

\subsection{Comparing Attribution Quality}

We subsequently aim to understand whether Split IG leads to improved attribution scores compared with those of standard IG. A significant limitation in doing so is the lack of ground truth for attribution, making it difficult to assess and compare the quality of attribution scores.
Proposed metrics for evaluating attribution methods aim to  quantify the following desiderata~\cite{yeh2019fidelity}:
\begin{enumerate}
    \item \textbf{Model Faithfulness} - Attribution scores should be faithful to the model being explained and accurately reflect the importance or contribution of each feature.
    \item \textbf{Stability} - Perturbations that do not change model output substantially, such as additive noise, should  not change the attributions substantially.
\end{enumerate}

To quantify model faithfulness, proposed metrics generally involve ablating parts of the input and comparing the change in model output with the computed attribution scores~\cite{petsiuk2018rise}. A simple approach is to use the ranking of input features based on attribution scores: removing the top features should result in a larger drop in model output, or alternatively, removing the bottom features should result in a limited drop in the output \cite{sturmfels2020visualizing}. The ABPC metric~\cite{samek2016evaluating} combines these ideas by computing two curves that correspond to the model output when ablating the top-ranked features (pixels) and bottom-ranked features respectively. The area between these curves represents the model faithfulness. An area of 0 indicates that ablating the top-ranked and bottom-ranked features has the same impact. On the other hand, larger values suggest that the ranking is more aligned with importance. We compute the ABPC metric for LeftIG, RightIG, and full IG,  estimating the area by ablating pixels with a zero baseline in increments of 10\% of the total number of pixels.

To quantify stability, we measure the sensitivity of each attribution, computed as the maximum change in attribution within a bounded space around the original input \cite{yeh2019fidelity}. We normalize by the original attribution magnitude to warrant fair comparison between attributions with different magnitudes. Formally, for input $x$, attribution functional $\Phi$,  and radius $r$, we compute the sensitivity as follows:
\[Sens(\Phi, x) = \frac{\max_{||\delta||_{\inf} < r} ||\Phi(x + \delta) - \Phi(x)||_2}{||\Phi(x)||_2} \]

We take 10 random samples within the L$_\infty$-bounded space around each normalized image with $r = 0.05$ to obtain a Monte Carlo approximation of the maximum change.
\begin{table}[H]
\caption{\textbf{Average area between perturbation curves for left, right, and full Integrated Gradients ($\psi = 0.9$)}. Left IG shows substantially higher ABPC than right IG and full IG.}
\vspace*{2mm}
\label{table:AreaBetwPertCurves}
\begin{tabular}{ c||p{1.6cm}|p{1.6cm}|p{1.7cm} }
 Model & LeftIG & RightIG & Full IG \\
 \hline
 Inception-v3   &  \textbf{0.2837}    & 0.1570 &   0.2486\\
 ResNet-50 &   \textbf{0.1708}  & 0.0879   & 0.1464\\
 VGG-19 & \textbf{0.1417} & 0.0810 &  0.1282\\
\end{tabular}
\end{table}
The results for these metrics are shown in Tables \ref{table:AreaBetwPertCurves} and \ref{table:Sens}. We observe that LeftIG consistently exhibits a substantially higher area between perturbation curves than RightIG, with full IG consistently being between the two. This suggests that the feature ranking based on unsaturated gradients is more faithful to the model than that of saturated gradients. Moreover, considering only the unsaturated regions leads to improved area between the perturbation curves compared to full integrated gradients.
\begin{table}[H]
\caption{\textbf{Average sensitivity of left, right and full Integrated Gradients for $\psi = 0.9$}. Left IG consistently shows substantially lower sensitivity than right IG and full IG.}
\vspace*{2mm}
\label{table:Sens}
\begin{tabular}{ c||p{1.6cm}|p{1.6cm}|p{1.7cm} }
 Model & LeftIG & RightIG & Full IG \\
 \hline
 Inception-v3   & \textbf{0.5341} & 0.6706 &   0.5711\\
 ResNet-50 &   \textbf{0.7706}  & 0.9145   &0.8074\\
 VGG-19 & \textbf{0.5970} & 0.8955 &  0.7122\\
\end{tabular}
\end{table}
We also observe consistently lower sensitivity across model architectures for LeftIG than RightIG, suggesting that the unsaturated gradients are less sensitive to input perturbations. These results support the conclusion that Split IG provides attribution scores that are more faithful to the model and less sensitive to noise than standard integrated gradients.

\subsection{Inclusion of Softmax}

A key choice in the preceding experiments evaluating saturation is attributing with respect to the model's logits, without including the final softmax layer. This distinction is particularly salient when investigating saturation, since the squashing property of softmax can cause further saturation of the model output \cite{sundararajan2016gradients}. Attribution methods such as DeepLift recommend computing attributions with outputs prior to this non-linearity to avoid the effects of this phenomenon \cite{shrikumar2017learning}. 

To better understand the change in IG results induced by the softmax function $S$, we compare the gradient of the logit function $F: R^i \rightarrow R^n$ for target class $t$ with and without applying $S$:

\[\frac{\partial S_t(F(x))}{\partial x} = \sum_{i = 1}^n \frac{\partial S_t(F(x))}{\partial F_i(x)} \cdot \frac{\partial F_i(x)}{\partial x} \]

\[ = \sum_{i = 1}^n S_t(F(x)) \cdot (\delta_{it} - S_i(F(x))) \cdot \frac{\partial F_i(x)}{\partial x}\]

Ignoring the terms where $i \neq t$, which incorporate gradients of logits other than the target logit, the remaining term is 

\[S_t(F(x)) \cdot (1 - S_t(F(x))) \cdot \frac{\partial F_t(x)}{\partial x}\]
In this term, the target logit gradient is multiplied by $S_t(F(x))\cdot(1 - S_t(F(x))$, which approaches 0 as $S_t(F(x)) \rightarrow 1$. In the context of IG, this term can help to mitigate the magnitude of saturated gradients in some cases, particularly when softmax output score approaches 1, but this depends on the value of the softmax output and would not mitigate saturation effects generally.

\section{Discussion and Takeaways}

The first key takeaway is that IG may be dominated by gradients in the saturated area, where model output does not change substantially, which counters an intuitive understanding of the method. Restricting the integral path to a smaller region that covers the substantial change in output still approximately satisfies the completeness property and often results in substantially different attributions. With standard ImageNet models, these attributions are less sensitive to noise and more faithful to the model, which are desirable properties of attribution results. On the other hand, the integrated gradients of saturated areas seem to be noisier and substantially less faithful to the model, which supports claims in prior work regarding the limitations of utilizing saturated gradient information for attribution \cite{sundararajan2016gradients}.

Based on these results, we recommend that users of IG also look at the model output along the integral path and visualize the gradients along this path as shown in Figure~\ref{fig:concept}, instead of relying on the overall integrated gradients attribution. This can give insights into the extent of saturation and how the gradients change along the path. In cases where saturation is present and the gradients in the saturated region contribute substantially to the overall integrated gradients, we suggest that users also compare the attributions with Split IG attributions, which may provide more useful attributions.

Finally, users should also be mindful of the distinction between computing IG for model output before or after a final softmax layer. Computing IG for a softmax output can either mute gradients in saturated regions when the saturated output approaches 1, or amplify it when the output approaches a smaller value.

\section{Related Work and Future Directions}
A variety of post-hoc model interpretability methods have been proposed for feature importance attribution. Two broad categories are gradient-based and perturbation-based methods~\cite{grun2016taxonomy}. Gradient-based methods make use of input gradients to estimate feature importance. Beside IG, this category includes Saliency~\cite{simonyan2013deep}, Expected Gradients~\cite{lundberg2017unified}, GradCAM~\cite{selvaraju2017grad}, and Guided Backpropagation~\cite{springenberg2014striving}.
Perturbation-based methods such as feature ablation and occlusion~\cite{zeiler2014visualizing} rely solely on model evaluations on perturbations of the input to estimate feature importance.

Given the popularity of IG due to its desirable theoretical properties, several variants and extensions to it have been explored. For example, Conductance~\cite{DBLP:journals/corr/abs-1805-12233} extends IG to quantify the importance of an intermediate neuron as opposed to an input feature when attributing the model output. 
Other extensions have focused on improving IG for certain model types and domains. Generalized Integrated Gradients (GIG) extends IG to be applicable to certain classes of discontinuous functions for models in domains such as financial services \cite{merrill2019generalized}. XRAI is a variant specifically for image attribution which combines IG with image segmentation for attribution to image regions rather than pixels \cite{kapishnikov2019xrai}.

Further work is needed to examine the effect of baseline choices~\cite{sturmfels2020visualizing} and model architecture on saturation as well as understanding the impacts of saturation on other gradient-based attribution methods and variants of IG. Metrics for evaluating attribution quality are limited due to the lack of a ground-truth. Nevertheless, further evaluation of existing metrics across datasets and architectures would be helpful to assess the generality of saturation effect.
Of particular interest is evaluating whether removing saturated areas in IG substantially improves attributions beyond standard image classification tasks and how these attribution results compare with those of other gradient-based and perturbation-based methods. 

\section{Conclusion}

We explore the effects of saturation on integrated gradients and find that regions along the path where the output changes minimally can have substantial effects on the overall integrated gradients. We propose a variant of integrated gradients that discards these regions. This variant captures gradients prior to saturation and exhibits improved faithfulness to the model and less sensitivity to noise than standard integrated gradients for ImageNet classification networks. The results suggest that users applying integrated gradients should (1) pay attention to the extent of contributions of gradients in saturated regions to the standard IG attribution and (2) compare attribution quality between the unsaturated and saturated regions to obtain better insight into integrated gradient results. 

\section{Aknowledgements}
We would like to thank Aniruddh Raghu for helpful feedback on an initial draft of this work.

\bibliography{example_paper}
\bibliographystyle{icml2020}

\onecolumn

\section*{Appendix}
Here we present further examples of computing the attribution for Inception-v3 using Split Integrated Gradients.
The first column shows the input images, with class labels listed in the titles.
The second and thrid columns depict the attribution results computed respectively by LeftIG, RightIG, with $\psi = 0.9$.
The fourth column depicts the attribution results computed using standard IG.
Green and red respectively indicate positive and negative attribution.
Generally, the output of LeftIG is less noisy than RightIG and standard IG and better matches the object ground truth.

\begin{figure*}[h!]
    \centering
    \includegraphics[width=0.8\linewidth]{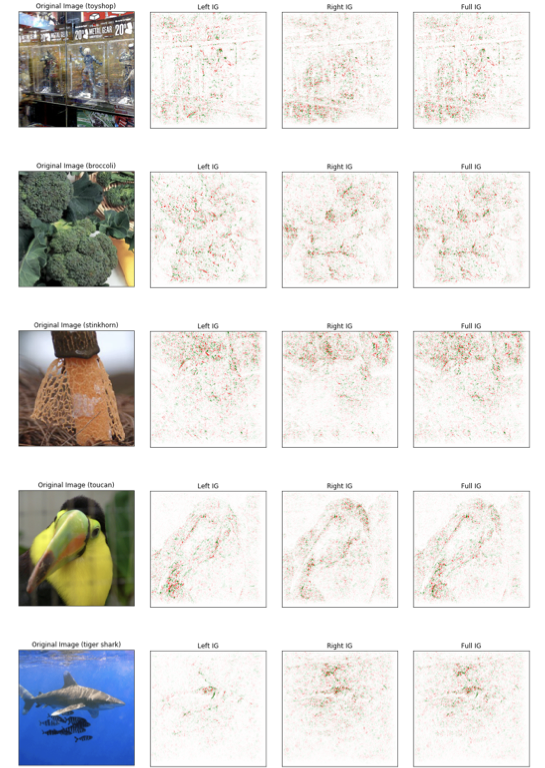}
    \label{fig:supp1}
\end{figure*}

\begin{figure*}[ht]
    \centering
    \includegraphics[width=0.8\linewidth]{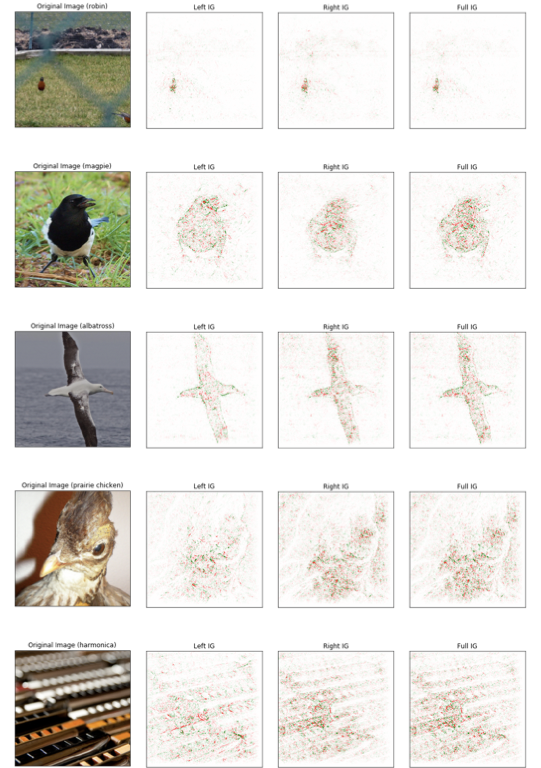}
    \label{fig:supp2}
\end{figure*}

\end{document}